\documentclass[sigconf]{acmart}

\usepackage{booktabs}
\usepackage{caption}
\usepackage{subfig}
\usepackage{times}
\usepackage{graphicx}
\usepackage{amssymb,amsmath,amsfonts}
\usepackage{multirow}
\usepackage{comment}
\usepackage{setspace}
\usepackage{afterpage}
\usepackage{url}
\usepackage[noend]{algpseudocode}
\usepackage{fancyvrb}
\usepackage{tabularx}
\usepackage{enumitem}

\setcopyright{none}
\acmConference[SysML'18]{SysML Conference}{February 2018}{Stanford, CA, USA}
\acmYear{2018}
\acmDOI{}
\acmISBN{}

\begin{document}
\title{Precision and Recall for Range-Based Anomaly Detection}

\author{Tae Jun Lee}
\authornote{The work was done while a Brown student.}
\affiliation{
  \institution{Microsoft}
%  \city{Redmond}
%  \state{WA}
}
%\email{tae_jun_lee@alumni.brown.edu}

\author{Justin Gottschlich, Nesime Tatbul}
\affiliation{
  \institution{Intel Labs}
%  \city{Santa Clara}
%  \state{CA}
}
%\email{{justin.gottschlich, nesime.tatbul}@intel.com}

\author{Eric Metcalf, Stan Zdonik}
\affiliation{
  \institution{Brown University}
%  \city{Providence}
%  \state{RI}
}
%\email{{emetcalf, sbz}@cs.brown.edu}

\renewcommand{\shortauthors}{T. J. Lee, J. Gottschlich, N. Tatbul, E. Metcalf, S. Zdonik}

\begin{abstract}

%Anomaly detection is the process of distinguishing anomalous events from normal ones.
%, has a nearly unlimited range of use and can add significant value to the domain in which it is applied. 
Classical anomaly detection is principally concerned with \emph{point-based anomalies}, anomalies that occur at a single data point. 
%While point-based anomalies are useful, most real-world anomalies are \emph{range-based}, meaning they occur over a range of time rather than at a single point. 
In this paper, we present a new mathematical model to express \emph{range-based anomalies}, anomalies that occur over a range (or period) of time.
\end{abstract}

\maketitle

\begin{abstract}

%Anomaly detection is the process of distinguishing anomalous events from normal ones.
%, has a nearly unlimited range of use and can add significant value to the domain in which it is applied. 
Classical anomaly detection is principally concerned with \emph{point-based anomalies}, anomalies that occur at a single data point. 
%While point-based anomalies are useful, most real-world anomalies are \emph{range-based}, meaning they occur over a range of time rather than at a single point. 
In this paper, we present a new mathematical model to express \emph{range-based anomalies}, anomalies that occur over a range (or period) of time.
\end{abstract}

\section{Introduction} \label{sec:introduction}

\emph{Anomaly detection} (AD) seeks to identify atypical events. Anomalies tend to be domain or problem specific, and many occur over a period of time. We refer to such events as \emph{range-based anomalies}, as they occur over a range (or period) of time\footnote{Range-based anomalies are a specific type of collective anomalies \cite{chandola:2009:ad-survey}. Moreover, range-based anomalies are similar, but not identical, to sequence anomalies \cite{warrender:1999:sequence}.}.
%For example, in medical applications, one might wish to represent each unique disease or illness as its own anomaly. Yet, if researchers were interested specifically in cancer detection, they might define each advancing stage of cancer as its own anomalous type. In the cyber-security space, each individually known vulnerability might be classified as its own anomaly as a means to track frequency. Yet, in network intrusion applications, each type of cyber-attack might be grouped together to help extrapolate analytics such as commonality, rarity, and emergency. In software correctness systems, one might wish to represent each type of race condition in parallel inter-leavings as its own anomaly. % while other performance-oriented systems may want to categorize anomalies by the parallel asymptotic computational complexity of the synchronization type used.
%Although the problems in each of the prior examples are different, they all have one thing in common. Each of their anomalies are events that occur over a period of time rather than at a fixed point. Disease and illness manifest gradually~\cite{kourou:cancer:2015}. Cyber-attacks are almost always multi-step processes~\cite{aleroud:2012:CAD}. Software correctness and performance bugs generally emerge from the execution of several serial operations in a precise sequence~\cite{mejbah:2016:act}. 
Therefore, it is critical that the accuracy measures for anomalies, and the systems detecting them, capture events that occur over a range of time. Unfortunately, classical metrics for anomaly detection were designed to handle only fixed-point anomalies~\cite{aggarwal:2013:outlier}. 
%The emphasis of this paper is to address this problem.
%\subsection{Recall and Precision}
%Perhaps the two most common and fundamental ways of capturing the accuracy of an anomaly detection system (ADS) is by using recall and precision.
An AD algorithm behaves much like a pattern recognition and binary classification algorithm: it recognizes certain patterns in its input and classifies them as either normal or anomalous. For this class of algorithms, {\em Recall} and {\em Precision} are widely used for evaluating the accuracy of the result. They are formally defined as in Equations~\ref{eq:recall} and ~\ref{eq:precision}, where \(\mathit{TP}\) denotes true positives, \(\mathit{FP}\) denotes false positives, and \(\mathit{FN}\) denotes false negatives.

\vspace{-0.2cm}
\begin{footnotesize}
\begin{align}
Recall & = TP \div (TP + FN) \label{eq:recall} \\
Precision & = TP \div (TP + FP) \label{eq:precision}
\end{align}
\end{footnotesize}
\vspace{-0.4cm}

%Informally, \emph{Recall} is the rate at which a system can identify anomalies without mispredicting anomalous events, while \emph{Precision} is the rate a system can identify anomalies without mispredicting non-anomalous ones. In this sense, {\em Recall} and {\em Precision} are opposites. This characterization proves useful when they are combined, such as is done when generating an $F_1$ or $F_{\beta}$ score, because such combinations help gauge the quality for both anomalous and non-anomalous predictions. 

%Unfortunately, these classical definitions of recall and precision were intended for \emph{point-based} anomalies (i.e., anomalies that occur precisely at a single point in time). 
\noindent
While useful for point-based anomalies, classical recall and precision suffer from their inability to capture, and bias, classification correctness for domain-specific time-series anomalies. Because of this, many time-series AD systems' accuracy are being misrepresented, as point-based recall and precision are used to measure their effectiveness \cite{singh-ijcnn17}. Furthermore, the need to accurately identify time-series anomalies is growing due to the explosion of streaming and real-time systems~\cite{twitter:2015:anom, malhotra:2015:esann, guha:2016:icml, ahmad:anomaly:2017, greenhouse-sysml18}. To address this, we redefine recall and precision to encompass range-based anomalies. Unlike prior work~\cite{lavin:anomaly:2015, ahmad:anomaly:2017}, our mathematical definitions are a superset of the classical definitions, enabling our system to subsume point-based anomalies. Moreover, our system is broadly generalizable, providing specialization functions to control a domain's bias along a multi-dimensional axis that is necessary to accommodate the needs of specific domains.

%The goal of this work is to design a mathematical model which can be used to evaluate, rank, and compare results of anomaly detection algorithms. We particularly focus on algorithms that target range-based anomalies (a.k.a., collective anomalies \cite{ad-survey}) over sequential data such as time-series. Thus, each anomaly can span a contiguous interval of time.
%In contrast with ``point-based anomalies", we call these ``range-based anomalies".
\begin{comment}
Our key technical contributions, in compressed form, include:~\footnote{Empirical data has been omitted to meet the venue's compressed format.}
\begin{enumerate}
\item{Novel formal definitions of recall and precision for range-based anomaly detection that both subsume those formerly defined for point-based anomaly detection as well as being customizable to a rich set of application domains.}
\end{enumerate}
\end{comment}
In this short paper, we present novel formal definitions of recall and precision for range-based anomaly detection that both subsume those formerly defined for point-based anomaly detection as well as being customizable to a rich set of application domains. Empirical data has been omitted to meet the venue's compressed format.
%\item{Empirical analysis of our new recall and precision model in comparison to the classical model as well as a recent scoring model provided by the Numenta Anomaly Benchmark~\cite{lavin:anomaly:2015}, across several different datasets.}
%\item{Algebraic analysis of our accuracy model based on Allen's interval algebra~\cite{allen}.}
%\end{enumerate}

\begin{figure*}[t]
\begin{center}
\subfloat[Overlap Size]{
\begin{minipage}[h]{0.95\columnwidth}
\begin{footnotesize}
\begin{algorithmic}
\Function{$\omega$}{\texttt{AnomalyRange}, \texttt{OverlapSet}, $\delta$}
\State $\texttt{MyValue} \gets 0$
\State $\texttt{MaxValue} \gets 0$
\State $\texttt{AnomalyLength} \gets \texttt{length(AnomalyRange)}$
\For{$\texttt{i} \gets 1, \texttt{AnomalyLength}$}
\State $\texttt{Bias} \gets \delta(\texttt{i}, \texttt{AnomalyLength})$
\State $\texttt{MaxValue} \gets \texttt{MaxValue} + \texttt{Bias}$
\If{\texttt{AnomalyRange}[i] in \texttt{OverlapSet}}
\State $\texttt{MyValue} \gets \texttt{MyValue} + \texttt{Bias}$
\EndIf
\EndFor
\State \Return $\texttt{MyValue}/\texttt{MaxValue}$
\EndFunction
\end{algorithmic}
\end{footnotesize}
\label{fig:omega}
\end{minipage}
}
%\hfill
\subfloat[Positional Bias]{
\begin{minipage}[h]{0.7\columnwidth}
\begin{footnotesize}
\begin{algorithmic}
\State // Flat positional bias
\Function{$\delta$}{\texttt{i}, \texttt{AnomalyLength}}
\State \Return 1
\EndFunction
\end{algorithmic}
\vspace{0.2cm}
\begin{algorithmic}
\State // Front-end positional bias
\Function{$\delta$}{\texttt{i}, \texttt{AnomalyLength}}
\State \Return \texttt{AnomalyLength} - \texttt{i} + 1
\EndFunction
\end{algorithmic}
\vspace{0.2cm}
\begin{algorithmic}
\State // Tail-end positional bias
\Function{$\delta$}{\texttt{i}, \texttt{AnomalyLength}}
\State \Return \texttt{i}
\EndFunction
\end{algorithmic}
\end{footnotesize}
\label{fig:delta}
\end{minipage}
}
\caption{Example Functions for $\omega()$ and $\delta()$}
\label{fig:example-func}
\end{center}
\end{figure*}
\vspace{-0.1cm}

\section{Range-based Recall} \label{sec:recall}

Classical {\em Recall} rewards an AD system when anomalies are successfully identified (i.e., TP) and penalizes it when they are not (i.e., FN). It is computed by counting the number of anomalous points successfully predicted and then dividing that number by the total number of anomalous points. However, it is not sensitive to domains where a single anomaly can be represented as a range of contiguous points. In this section, we propose a new way to compute recall for such range-based anomalies. Table \ref{tab:notation} summarizes our notation.

\begin{table}[t]
\small
\caption{Notation}
\begin{center}
\begin{tabular}{|c|l|}
\hline
{\bf Notation} & {\bf Description} \\
\hline
\hline
$R$ & set of real anomaly ranges \\
\hline
$R_i$ & the $i^{th}$ real anomaly range \\
\hline
$P$ & set of predicted anomaly ranges \\
\hline
$P_j$ & the $j^{th}$ predicted anomaly range \\
\hline
$N_r$ & number of real anomaly ranges \\
\hline
$N_p$ & number of predicted anomaly ranges \\
\hline
$\alpha$ & relative weight of existence reward \\
\hline
$\beta$ & relative weight of overlap reward \\
\hline
$\gamma()$ & overlap cardinality function \\
\hline
$\omega()$ & overlap size function \\
\hline
$\delta()$ & positional bias function \\
\hline
\end{tabular}
\label{tab:notation}
\end{center}
\end{table}

Given a set of real anomaly ranges $R=\{R_1, .., R_{N_r}\}$ and a set of predicted anomaly ranges $P=\{P_1, .., P_{N_p}\}$, our $Recall_{T}(R, P)$ formulation iterates over the set of all real anomaly ranges ($R$), computing a recall score for each real anomaly range ($R_i \in R$) and adding them up into a total recall score. This total score is then divided by the total number of real anomalies ($N_r$) to obtain an average recall score for the whole time-series.

\begin{footnotesize}
\begin{align}
Recall_{T}(R, P) & = \frac{\sum_{i=1}^{N_r}{Recall_{T}(R_i, P)}}{N_r}
\label{eq:recall1}
\end{align}
\end{footnotesize}

\noindent
When computing the recall score $Recall_{T}(R_i,P)$ for a single real anomaly range $R_i$, we take the following aspects into account:

\begin{itemize}
\item \emph{Existence}: Identifying an anomaly (even by a single point in $R_i$) may be valuable in some application domains.
\item \emph{Size}: The larger the size of the correctly predicted portion of $R_i$, the higher the recall score will likely be.
\item \emph{Position}: In some cases, not only size, but also the relative position of the correctly predicted portion of $R_i$ may be important to the application (e.g., early and late biases).
\item \emph{Cardinality}: Detecting $R_i$ with a single predicted anomaly range $P_j \in P$ may be more valuable to an application than doing so with multiple different ranges in $P$.
\end{itemize}

\noindent
We capture these aspects as a sum of two reward terms weighted by $\alpha$ and $\beta$, respectively, where $0 \leq \alpha, \beta \leq 1$ and $\alpha+\beta=1$. $\alpha$ represents the relative importance of rewarding {\em existence}, whereas $\beta$ represents the relative importance of rewarding {\em size}, {\em position}, and {\em cardinality}, which all stem from the overlap between $R_i$ and the set of all predicted anomaly ranges ($P_i \in P$). 

\vspace{-0.2cm}
\begin{footnotesize}
\begin{align}
Recall_{T}(R_i, P) & = \alpha \times ExistenceReward(R_i, P) \nonumber \\
& + \beta \times OverlapReward(R_i, P)
\label{eq:recall2}
\end{align}
\end{footnotesize}
\vspace{-0.2cm}

\noindent
If anomaly range $R_i$ is identified (i.e., $\vert R_i \,\, \cap \,\, P_j \vert \geq 1$ across all $P_j \in P$), then an existence reward of $1$ is earned.

\vspace{-0.2cm}
\begin{footnotesize}
\begin{align}
\hspace{-0.2cm}
ExistenceReward(R_i, P) & =
\begin{cases}
1 & \text{\hspace{-0.1cm}, if} \sum_{j=1}^{N_p}{\vert R_i \cap P_j \vert \geq 1} \\
0 & \text{\hspace{-0.1cm}, otherwise}
\end{cases}
\label{eq:existence}
\end{align}
\end{footnotesize}

\noindent
Additionally, an overlap reward, dependent upon three application-defined functions $0 \leq \gamma() \leq 1$, $0 \leq \omega() \leq 1$, and $\delta() \geq 1$, can be earned. These functions capture the {\em cardinality} ($\gamma$), {\em size} ($\omega$), and {\em position} ($\delta$) of the overlap. The cardinality term serves as a scaling factor for the rewards earned from size and position of the overlap.

\vspace{-0.2cm}
\begin{footnotesize}
\begin{align}
OverlapReward(R_i, P) & = CardinalityFactor(R_i, P) \nonumber \\
& \times \sum_{j=1}^{N_p}{\omega(R_i, R_i \cap P_j, \delta)}
\label{eq:overlap}
\end{align}
\end{footnotesize}
\vspace{-0.2cm}

\noindent
The cardinality factor is largest (i.e., 1), when $R_i$ overlaps with at most one predicted anomaly range (i.e., it is identified by a single prediction range). Otherwise, it receives a value $0 \leq \gamma() \leq 1$ defined by the application.

\vspace{-0.3cm}
\begin{footnotesize}
\begin{align}
\hspace{-0.3cm}
CardinalityFactor(R_i, P) & =
\begin{cases}
1 & \parbox[t]{1.8cm}{\hspace{-0.2cm}, if $R_i$ overlaps with at most one $P_j \in P$} \\
\gamma(R_i, P) & \text{\hspace{-0.2cm}, otherwise}
\end{cases}
\label{eq:cardinality}
\end{align}
\end{footnotesize}
\vspace{-0.1cm}

\noindent
The $Recall_{T}$ constants ($\alpha$ and $\beta$) and functions ($\gamma()$, $\omega()$, and $\delta()$) are tunable according to the needs of the application. Next, we illustrate how they can be customized with examples.

The cardinality factor should generally be inversely proportional to $Card(R_i)$, i.e., the number of distinct prediction ranges that a real anomaly range $R_i$ overlaps. For example, $\gamma(R_i, P)$ can simply be set to $1/Card(R_i)$.

Figure \ref{fig:omega} provides an example for the $\omega()$ function for size, which can be used with many different $\delta()$ functions for positional bias as shown in Figure \ref{fig:delta}. If all index positions are equally important, then the flat bias function should be used. If earlier ones are more important than later ones (e.g., early cancer detection \cite{kourou:cancer:2015}, real-time apps \cite{ahmad:anomaly:2017}), then the front-end bias function should be used. Finally, if later index positions are more important (e.g., delayed response in robotic defense), then the tail-end bias function should be used. 

Our recall formula for range-based anomalies subsumes the classical one for point-based anomalies (i.e., $Recall_{T} \equiv Recall$) when:
\begin{enumerate}[nosep,label=(\roman*)]
\item all $R_i \in R$ and $P_j \in P$ are represented as single-point ranges (e.g., range $[1,3]$ represented as $[1,1],[2,2],[3,3]$), and
\item $\alpha=0, \beta=1, \gamma()=1$, $\omega()$ is as in Figure \ref{fig:omega}, and $\delta()$ returns flat positional bias as in Figure \ref{fig:delta}.
\end{enumerate}

\vspace{-0.06in}
\section{Range-based Precision} \label{sec:precision}

Classical $Precision$ is computed by counting the number of successful prediction points (i.e., TP) in proportion to the total number of prediction points (i.e., TP+FP). The key difference between {\em Precision} and {\em Recall} is that {\em Precision} penalizes FPs. In this section, we extend classical precision to handle range-based anomalies. Our formulation follows a similar structure as $Recall_{T}$.

Given a set of real anomaly ranges $R=\{R_1, .., R_{N_r}\}$ and a set of predicted anomaly ranges $P=\{P_1, .., P_{N_p}\}$, $Precision_{T}(R, P)$ iterates over the set of predicted anomaly ranges ($P$), computing a precision score for each range ($P_i \in P$) and then sums them. This sum is then divided by the total number of predicted anomalies ($N_p$), averaging the score for the whole time-series.

\vspace{-0.2cm}
\begin{footnotesize}
\begin{align}
Precision_{T}(R, P) & = \frac{\sum_{i=1}^{N_p}{Precision_{T}(R, P_i)}}{N_p}
\label{eq:precision1}
\end{align}
\end{footnotesize}
\vspace{-0.2cm}

\noindent
When computing $Precision_{T}(R,P_i)$ for a single predicted anomaly range $P_i$, there is no need for an {\em existence} reward, because precision by definition emphasizes prediction quality, and existence by itself is too low a bar for judging the quality of a prediction. This removes the need for $\alpha$ and $\beta$ constants. Therefore:

\vspace{-0.2cm}
\begin{footnotesize}
\begin{align}
Precision_{T}(R, P_i) & = CardinalityFactor(P_i, R) \nonumber \\
& \times \sum_{j=1}^{N_r}{\omega(P_i, P_i \cap R_j, \delta)}
\label{eq:precision2}
\end{align}
\end{footnotesize}
\vspace{-0.2cm}

\noindent
$\gamma()$, $\omega()$, and $\delta()$ are customizable as before. Furthermore, $Precision_{T}$ $\equiv Precision$ under the same settings as in Section \ref{sec:recall} (except $\alpha$ and $\beta$ are not needed). Note that, while $\delta()$ provides a potential knob for positional bias, we believe that in many domains a flat bias function will suffice for $Precision_{T}$, as an $FP$ is typically considered uniformly bad wherever it appears in a prediction range.

%It is important to notice that the positional bias function ($\delta()$) used in $Precision_{T}$ refers to relative index positions in a predicted anomaly range, not in a real anomaly range (as it was the case in $Recall_{T}$). Furthermore, "front-end" of a predicted anomaly range may not necessarily correspond to the "front-end" of a real anomaly range; it depends on how the two ranges are aligned with respect to each other. For example, in Figure \ref{fig:intervals}, the front-end of the first predicted anomaly range ($P_1$) lies near the tail-end of the first real anomaly range ($R_1$); whereas the front-end of the second predicted anomaly range ($P_2$) lies near the front-end of the second real anomaly range ($R_2$). The $\omega()$ and $\delta()$ functions of $Precision_{T}$ must be properly customized according to what the application considers as "positional bias" (if any) for a predicted range in terms of the quality/exactness of that prediction, and might be different from the ones defined for $Recall_{T}$.

% todo: need a paragraph here for why this basic precision is not sufficient to motivate the advanced version that follows...

\vspace{-0.06in}
\section{Conclusion} \label{sec:conclusion}

In this paper, we note that traditional recall and precision were invented for point-based analysis. In range-based anomaly detection, anomalies are not necessarily single points, but are, in many cases, ranges. In response, we offered new recall and precision definitions that take ranges into account. 

%Things like overlap of the predicted vs. the real ranges are accounted for in our definitions. Moreover, we acknowledge that latency is an important characteristic of any anomaly detection algorithm, and our approach allows users to supply bias functions that weigh front-end and back-end overlap differently. Finally, we show that by setting parameters correctly, our model can be made to behave like the classical one.

\vspace{0.05in}
\noindent
{\bf Acknowledgments.}
This research has been funded in part by Intel.

\newpage

\bibliographystyle{ACM-Reference-Format}
\bibliography{anomaly}

\end{document}